\documentclass[journal]{IEEEtran}

\ifCLASSINFOpdf
\else
\fi

\usepackage{hyperref}        
\usepackage{url}             
\usepackage{booktabs} 
\usepackage{amsfonts} 
\usepackage{nicefrac}
\usepackage{microtype}
\usepackage{bookmark}
\usepackage{array}
\usepackage{graphicx}
\usepackage{amsmath}
\usepackage{amsthm}
\usepackage{amsfonts} 
\usepackage{amssymb}
\usepackage{array}
\usepackage{algorithmic}
\usepackage{algorithm}
\usepackage{tabularx}
\usepackage{subfigure}
\usepackage{color}
\usepackage{cite}
\usepackage{cases}

\newtheorem{definition}{Definition}
\newtheorem{remark}{Remark}
\newtheorem{lemma}{Lemma}
\hypersetup{hidelinks}

\newcommand{\slw}{\color{black}}
\newcommand{\rv}{\color{black}}

\begin{document}

\title{Electricity Price Prediction for Energy Storage System Arbitrage: A Decision-focused Approach}

\author{Linwei Sang,
Yinliang Xu$^*$,~\IEEEmembership{Senior Member,~IEEE,}
Huan Long,~\IEEEmembership{Member,~IEEE,}
Qinran Hu,~\IEEEmembership{Senior Member,~IEEE,}
Hongbin Sun,~\IEEEmembership{Fellow,~IEEE}
\thanks{This work was partially supported by Shenzhen Science and Technology Program, Grant No. JCYJ20210324130811031, Guangdong Basic and Applied Basic Research Foundation, Grant No.2021A1515012450, and Tsinghua Shenzhen International Graduate School Interdisciplinary Research and Innovation Fund, Grant No. JC2021004. Paper no. TSG-01664-2021 (\textit{Corresponding Author: Yinliang Xu})}
\thanks{Linwei~Sang, Yinliang~Xu are with Tsinghua-Berkeley Shenzhen Institute, Tsinghua Shenzhen International Graduate School, Tsinghua University, Shenzhen, China. (E-mail: \url{sanglw21@mails.tsinghua.edu.cn}, \url{xu.yinliang@sz.tsinghua.edu.cn}.)}
\thanks{Huan~Long, Qinran~Hu are with the School of Electrical Engineering, Southeast University, Nanjing, China. (E-mail:\url{hlong@seu.edu.cn}, \url{qhu@seu.edu.cn})}
\thanks{H. Sun is with the Department of Electrical Engineering, State Key Laboratory of Power Systems, Tsinghua University, Beijing, China. (E-mail: \url{shb@tsinghua.edu.cn})}
}

\maketitle

\begin{abstract}
Electricity price prediction plays a vital role in energy storage system (ESS) management. Current prediction models focus on reducing prediction errors but overlook their impact on downstream decision-making. So this paper proposes a decision-focused electricity price prediction approach for ESS arbitrage to bridge the gap from the downstream optimization model to the prediction model. The decision-focused approach aims at utilizing the downstream arbitrage model for training prediction models. It measures the difference between actual decisions under {\slw the} predicted price and oracle decisions under {\slw the} true price, \textit{i.e.}, decision error, by regret, transforms it into the tractable surrogate regret, and then derives the gradients to predicted price for training prediction models. Based on the prediction and decision errors, this paper proposes the hybrid loss and corresponding stochastic gradient descent learning method to learn prediction models for prediction and decision accuracy. The case study verifies that the proposed approach can efficiently bring more economic benefits and {\slw reduce decision errors} by flattening the time distribution of prediction errors, compared to prediction models for only minimizing prediction errors.
\end{abstract}


\IEEEpeerreviewmaketitle

\section{Introduction}

\IEEEPARstart{D}{ue} to the high penetration of renewables and deregulation of the electricity market, electricity price becomes volatile \cite{Chitsaz2018, Peng2018}, and hence its accurate prediction is difficult. Electricity price prediction has widespread application in the smart grid, including the energy storage system (ESS) management and scheduling. The predicted price from prediction models is delivered to the downstream ESS scheduling model, making the optimal charging/discharging decisions to maximize its arbitrage benefits \cite{Arteaga2019}. The whole process follows the predict-then-optimize framework \cite{Elmachtoub2020}. Under this framework, various electricity price prediction methods have been proposed to improve prediction accuracy in recent literature.

{\rv 
Electricity price prediction is a fundamental technique in deregulated electricity markets \cite{Weron2014}. Conventional electricity price prediction focuses on the price prediction from the perspectives of prediction horizons and various prediction models \cite{Weron2014}. A two-stage electricity price forecast scheme is developed to predict electricity price spike in the first stage and continuous price in the second stage for improving prediction accuracy \cite{Shi2021}. Hybrid models are developed to improve prediction accuracy based on wavelet and LSTM networks \cite{Chang2019, Qiao2020}. Considering energy management and control, the multi-horizon electricity price forecasting models are proposed to improve prediction accuracy and detect the price spikes in \cite{Chitsaz2018}, revealing its economic value. The holistic approach is developed to recover the energy market structure and predict the nodal market electricity price \cite{Radovanovic2019}. Price interval prediction has come forth in recent years. In \cite{Wan2017}, a Pareto optimal prediction interval construction approach is proposed to predict the electricity price intervals, achieving the interval's reliability and sharpness. Considering the dynamic price changes from the neighboring regions, the online learning approach provides rolling day-ahead price/interval prediction and 30-min prediction by perceiving the neighboring price fluctuation for improving prediction accuracy \cite{Xiao2021}. A reinforcement learning-based decision system is proposed to assist the selection of pricing plans for minimizing the end-user payment and consumption payment \cite{Lu2021}. The above prediction-based works highlight the prediction accuracy of prediction models but overlook the prediction errors' direct impact on downstream optimization models.
}

{\rv
Due to high electricity price fluctuations, ESS can gain profits by charging at low prices and discharging at high prices \cite{Comello2019}. A storage scheduling algorithm is proposed for the joint arbitrage and operating reserve as merchant functions in \cite{Khani2019}, which can effectively utilize the storage for arbitrage benefits and reserve service. A non-complementary energy storage arbitrage model is developed by replacing the binary variables without jeopardizing practical viability  \cite{Shen2021}. A bi-level energy storage arbitrage model is constructed by considering the wind power and LMP smooth effect in \cite{Cui2018}, where the upper layer maximizes the arbitrage revenue and the lower layer simulates the market-clearing. Considering the uncertainty of wind and solar energy, a stochastic energy storage arbitrage model is developed to maximize its profit under the day-ahead and real-time market prices in \cite{Krishnamurthy2018}. An MPC-based ESS control policy is designed in \cite{Hashmi2020} to perform energy arbitrage and local power factor correction, where the electricity price is forecasted by auto-regression. In \cite{Cao2020}, deep reinforcement learning is developed for real-time ESS charging/discharging control based on the hybrid CNN and LSTM-based price prediction. The operation of the power system requires various ESS application scenarios, including synergies price arbitrage and fast frequency response \cite{Pusceddu2021}, simultaneous peak shaving and arbitrage \cite{Schneider2021}, and co-optimization of community energy systems \cite{Terlouw2019}. Considering the price prediction uncertainty, the MPC-based multi-objective scheduling is analyzed in \cite{Nair2021}. The above optimization-based works formulate the uncertainty of predicted price by sampling or distribution assumptions and seldom consider the optimization models' reverse impact on upstream prediction models. 
}

The above research indicates the gap between upstream prediction and downstream optimization models. The objective of conventional prediction models is usually to minimize the difference between the predicted price and true price, denoted by prediction error. In contrast, ESS focuses on maximizing arbitrage benefits from electricity price fluctuations. Maximizing benefits can be equivalent to minimizing the difference between the optimal actual decisions under the predicted price and optimal oracle decisions under the true price, denoted by the decision error. The decision error does not necessarily coincide with the prediction error. {\rv Recent years have seen increasing work on utilizing prediction considering optimization models. Embedding the differential economic dispatch within the neural network is developed in \cite{Donti2017} in the load prediction for the decision quality instead of prediction accuracy. And the embedding framework is further extended to the general economic dispatch in \cite{Lu2020, Han2021}. In \cite{Elmachtoub2020}, the smart ``predict, then optimize'' (SPO) loss is proposed to learn linear predictor parameters by linear programming to solve the shortest path and portfolio optimization problems.}

Inspired by the above works, we propose a decision-focused electricity price prediction approach to take advantage of the downstream ESS arbitrage model's decision error to train the prediction models. The prediction model target in the proposed approach focuses on minimizing prediction error in the prediction model and decision error in the ESS model. However, the decision error comes from the ESS arbitrage model and cannot apply to training the prediction models directly. To tackle this challenge, we first measure the decision error by regret, transform the regret into tractable surrogate regret, and derive the gradients of surrogate regret to the predicted price for training prediction models. These gradients instruct the prediction models to learn to reduce prediction model decision error. The prediction error is measured by the mean-square-error (MSE) metric. Based on MSE and surrogate regret, we design a hybrid loss function and propose a hybrid stochastic gradient descent (SGD) learning method of updating prediction models' parameters for prediction and decision accuracy. Finally, we apply the decision-focused learning approach in various prediction models to verify its effectiveness. 

{\rv To distinguish from previous works, the novelties of this paper are demonstrated in three aspects. 1) The proposed approach utilizes prediction and decision errors to train the price prediction model for ESS arbitrage, while Ref. \cite{Elmachtoub2020, Donti2017,Lu2020,Han2021} only considers the decision error. 2) Tractable surrogate regret is proposed for training a multi-layer neural network called residual neural network (ResNet) in the context of deep learning, while the predictor of Ref. \cite{Elmachtoub2020} with similar loss called the SPO loss is a linear model in the context of linear programming. 3) The hybrid loss design and learning method are proposed to improve decision accuracy and increase economic benefits, which are \textbf{first} applied in prediction for ESS arbitrage.}

{\slw In summary, the contribution of this paper is threefold:}

1) To the authors' best knowledge, this paper \textbf{first} takes advantage of both prediction error and decision error to learn the parameters of prediction models and proposes a decision-focused electricity price prediction approach for ESS arbitrage. Compared with previous prediction methods \cite{Shi2021, Chang2019, Qiao2020, Gonzalez2018, Diaz2019}, the proposed decision-focused approach pays attention to the prediction error's impact on the downstream optimization models, improving the decision accuracy under the predicted price.

2) Tractable surrogate regret is proposed to measure the decision error between the actual decision under the predicted price and the oracle decision under the true price. Its gradients to the predicted price are further derived, which is the key of this paper. Regret of decision under the predicted price is discontinuous and intractable. So this paper relaxes the feasible region of the original problem and derives a tractable upper bound of regret to measure the decision error, denoted by surrogate regret. Surrogate regret gradients to the predicted price are derived and delivered from the downstream optimization model to the upstream prediction model for reducing decision errors.

3) Based on the weighted sum of MSE and surrogate regret, this paper designs a hybrid loss function to measure prediction and decision errors. Then a hybrid SGD learning method for ESS arbitrage is proposed to train prediction models for improving prediction and decision accuracy based on hybrid loss. The case study indicates the proposed decision-focused approach can capture and predict the changing trends of electricity prices more accurately, which leads to the increase of decision accuracy and economic benefits.

The rest of this paper is organized as follows. Section II presents the conventional predict-then-optimize framework. Section III proposes the decision-focused prediction approach and the hybrid stochastic gradient descent method to learn prediction models. Section IV evaluates the effectiveness of the proposed decision-focused approach in the case study. Section V concludes this paper.

\section{Conventional Predict-then-optimize Framework}
\begin{figure}[ht]
  \centering
  \includegraphics[scale=1.1]{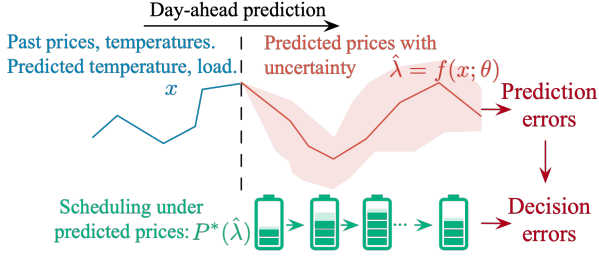}
  \caption{Day-ahead electricity prediction and ESS arbitrage under predict-then-optimize framework.}
  \label{Problemform}
\end{figure}

This section introduces the predict-then-optimize framework. Fig. \ref{Problemform} presents the day-ahead electricity price prediction and relevant ESS scheduling model, where prediction error is delivered from the upstream prediction model to the downstream ESS model, resulting in decision error.

\subsection{Upstream Electricity Price Prediction Model}
As shown in Fig. \ref{Problemform}, day-ahead electricity price prediction is the basis of energy scheduling. This part focuses on the price prediction model formulation, whose general form is as follows:
\begin{equation}
  \label{Prediction}
  \begin{aligned}
    \hat{\lambda} = f(x)
  \end{aligned}
\end{equation}
where $\hat{\lambda}$ is the prediction electricity price, $x$ is the input features, and $f(\cdot)$ denotes the prediction model. From input feature vector $x$ to output predicted price $\hat{\lambda}$, the whole process is composed of five procedures: i) data pre-processing, ii) feature engineering, iii) model selection, iv) training procedures, v) prediction objective.

\subsubsection{Data Pre-processing}
Data pre-processing transforms raw datasets into the prediction model input features and output datasets, including filling missing values, clearing outliers, and normalizing the datasets. 

\subsubsection{Feature Engineering}
Feature engineering selects and formulates the feature vectors from the processed dataset for inputting the prediction model. The electricity price is usually influenced by historical load and price, future load and temperature, and some calendar factors, including weekday/weekend, holiday effects, and day of the year. For more features input, we construct the squares of future load and temperature.

\subsubsection{Model Selection}
Prediction is generally formulated as a regression problem, which maps the input feature vectors into the continuous output prediction values. Various regression models are proposed for electricity price prediction in the smart grid, from conventional linear regression in the statistical domain to burgeoning neural networks in the deep learning domain. These models feature various strengthens and dataset requirements. 

In this work, we focus on comparing two kinds of prediction models with different representational capacities: i) the linear regression model {\rv \cite{Wang2020}} in Equation (\ref{Linear}); ii) the ResNet model {\rv \cite{Ren2021}} in Equation (\ref{ResNet}). It should be noted that the model representational capacity refers to the models' ability to fit variety of functions.
\begin{equation}
  \label{Linear}
  \begin{aligned}
    \hat{\lambda} = f(x; \theta) = \theta_x^{lr} x + \theta_b^{lr} \\ 
  \end{aligned}
\end{equation}
Linear regression model maps the input feature vector to predicted price by $\theta^{lr}_x$ and $\theta^{b}_x$ linearly, whose representational capacity is limited. 

\begin{equation}
  \label{ResNet}
  \begin{aligned}
     y_{l+1} = \sigma (\theta_{y,l}^{nn} y_l &+ \theta_{x,l}^{nn} x_l + \theta_{b,l}^{nn}) \quad l = 1, ..., N \\
     \text{where: }  x_1 &= x \\
    \hat{\lambda} &= y_{N} \\
  \end{aligned}
\end{equation}
{\rv where $x$ refers to the input features and $y_l$ refers to the output of the $l$ th layer in the neural network.}

While the ResNet model is a group of $N$ stacked linear regression and activation function mapping from input to output by a set of $\theta^{nn}_{y,\cdot}$, $\theta^{nn}_{x,\cdot}$, and $\theta^{nn}_{b,\cdot}$, whose representational capacity is large.

{\rv As shown in Equations \eqref{Linear} and \eqref{ResNet}, the parameter number of the linear model is much less than that of the ResNet model due to the multi-layers of the ResNet model, so its representational capacity is much smaller than ResNet models. The two models represent two typical prediction models, \textit{i.e.}, small and high representational prediction models. This paper applies the proposed decision-focused approach in different models to verify its generalization performance in different representational capacity models. In model selection, the ResNet model with high representational capacity is prioritized to capture the high stochasticity of electricity prices under massive training data and fierce price changing; the linear model with high representational capacity is prioritized under limited training data and mild price changing.}

\subsubsection{Training Procedures}
Training prediction model for better generalization follows these essential procedures: i) A subset of complete processed data is selected as a testing dataset randomly; ii) the remaining is further split into training and validation set randomly; iii) then we train the prediction model on the training set, evaluate its accuracy on the validation set, and tune model hyperparameters for high accuracy; iv) finally, we test the trained model on the testing set for model evaluation and comparison. 

\subsubsection{Objective of Prediction Models}
The objective of prediction models is to minimize the difference between {\slw the} predicted price $\hat{\lambda}$ and true price $\lambda$, described by MSE in Equation \eqref{MSE}.

\begin{equation}
  \label{MSE}
  \begin{aligned}
    L^{MSE}(\hat{\lambda}, \lambda) &= \frac{1}{T} \frac{||\hat{\lambda} - \lambda ||^2_2}{2}  = \frac{1}{T} \sum_{t=1}^T \frac{(\hat{\lambda}_t - \lambda_t)^2}{2} \\
  \end{aligned}
\end{equation}

Then the objective of prediction models is formulated in Equation (\ref{Targetofprediction}).
\begin{equation}
  \label{Targetofprediction}
  \begin{aligned}
    \min_{\theta} &L^{MSE}(\hat{\lambda}, \lambda) \\
    \text{s.t. } &\hat{\lambda} = f(x; \theta) \\
  \end{aligned}
\end{equation}

The gradients of MSE to the predicted price can further be derived as:

\begin{equation}
  \label{gradientsofMSE}
  \begin{aligned}
    \frac{\partial L^{MSE}(\hat{\lambda}, \lambda)}{\partial \hat{\lambda}} &= \frac{1}{T} \sum_{t=1}^T (\hat{\lambda}_t - \lambda_t) \\
    \text{s.t. } \hat{\lambda} &= f(x; \theta) \\
  \end{aligned}
\end{equation}
{\rv We note that the variables with the footnote $t$ refer to the time elements of the corresponding vector variables.} 

Based on the above, the training of prediction models usually utilizes the conventional SGD learning method. The model parameters are updated by way of batch updating \cite{Goodfellow2016} to improve the calculation efficiency.

\subsection{Downstream Price-based ESS Arbitrage Model}

ESS usually works at charging state when electricity price is low and at discharging state when electricity price is high for arbitrage from market. The operation of ESS usually follows the electricity price signals to maximize its benefits as:
{\slw
\begin{equation}
  \label{ESbigM}
  \begin{aligned}
    \max_{P \in \Phi_{ESS}} c(P; \lambda) = &\lambda^T P {\rv \Delta_t} = \sum_{t=1}^T \lambda_t P_t {\rv \Delta_t} \\ 
  \end{aligned}
\end{equation}}
where its objective is to maximize the net energy benefits by optimizing its operating power $P$ (MW) {\rv time interval $\Delta_t$} under market day-ahead electricity price $\lambda$ (\$/MWh), denoted by $c(P; \lambda)$. And we note that the variables with the footnote $t$ refer to the time elements of the corresponding vector variables. The feasible region $\Phi_{ESS}$ of decision variable $P$ is subject to a set of operation and technical constraints as follows:
\begin{equation}
  \label{Contrs1}
  \begin{aligned}
    P = P_{dis} - P_{ch} \\
  \end{aligned}
\end{equation}
Equation (\ref{Contrs1}) illustrates the operating power $P$ of ESS is composed of charging and discharging parts; $P_{dis}$ denotes the charging of ESS from grid, and $P_{ch}$ denotes the discharging of ESS to grid.

{\rv
\begin{equation}
  \label{Contrs2}
  \begin{aligned}  
    &E_t = E_{t-1} + \eta_{ch} P_{ch,t} \Delta_t - \frac{P_{dis,t}}{\eta_{dis}} \Delta_t \quad t=2,..,T\\
    &E_1 = E_{init} + \eta_{ch} P_{ch,1} \Delta_t - \frac{P_{dis,1}}{\eta_{dis}} \Delta_t \\
    &E_{min} \leq E_t \leq E_{max} \quad t=1,..,T \\ 
  \end{aligned}
\end{equation}}
{\rv where $E_t$ (kWh) refers to the stored energy in the ESS. Equation (\ref{Contrs2}) ensures $E_t$ in the ESS at time \textit{t} lies in an allowable range}, and $E_{min}$, $E_{max}$ refer to the minimum, maximum capacity of the battery system.
{\rv
\begin{equation}
  \label{Contrs3}
  \begin{aligned}
    & 0 \leq P_{dis,t} \leq \min\{P_{dis}^{max}, \eta_{dis}\frac{E_{t-1} - E_{min}}{\Delta_t} \} \quad t=1,...,T \\
    & 0 \leq P_{ch,t} \leq \min\{ P_{ch}^{max}, \frac{E_{max} - E_{t-1}}{\eta_{ch} \Delta_t} \} \quad t=1,...,T \\
    & 0 \leq P_{dis,t} \leq M \mu_{dis,t} \quad t=1,...,T \\
    & 0 \leq P_{ch,t} \leq M \mu_{ch,t} \quad t=1,...,T\\
    & \mu_{dis,t} + \mu_{ch,t} \leq 1 \quad t=1,...,T\\
  \end{aligned}
\end{equation}
}
Equation (\ref{Contrs3}) prevents the simultaneous charging and discharging of ESS by utilizing the big-M method; M is a large positive number. $\mu_{dis}$ and $\mu_{ch}$ are binary indicators of discharging and charging state, where $1$ means in the state and $0$ means the opposite. $P_{dis}^{max}$ and $P_{ch}^{max}$ are the maximum values of charging and discharging power. So the whole ESS arbitrage model is formulated as a mixed-integer linear program. {\rv We note that ignoring binary variables $\mu$ may lead to the suboptimality of the energy arbitrage model according to Ref. \cite{Shen2021}.}

Under the predict-then-optimize framework, the relationship between prediction and optimization is uni-direct, where the upstream prediction model delivers its price to the downstream optimization model explicitly. Simultaneously, the prediction error is also delivered to the downstream implicitly, which leads to the decision error. 

\section{Methodology}

This paper takes advantage of the downstream decision error to improve the prediction model to fill the gap from the optimization model to the prediction model. {\rv The prediction model should consider the prediction error and decision error in the training process.} This section presents the general decision-focused framework, measures the decision error by regret, formulates the tractable form of regret by surrogate regret, and finally proposes the hybrid SGD method learning for training prediction models.

\subsection{The Proposed Decision-focused Approach}

\begin{figure}[ht]
  \centering
  \includegraphics[scale=1]{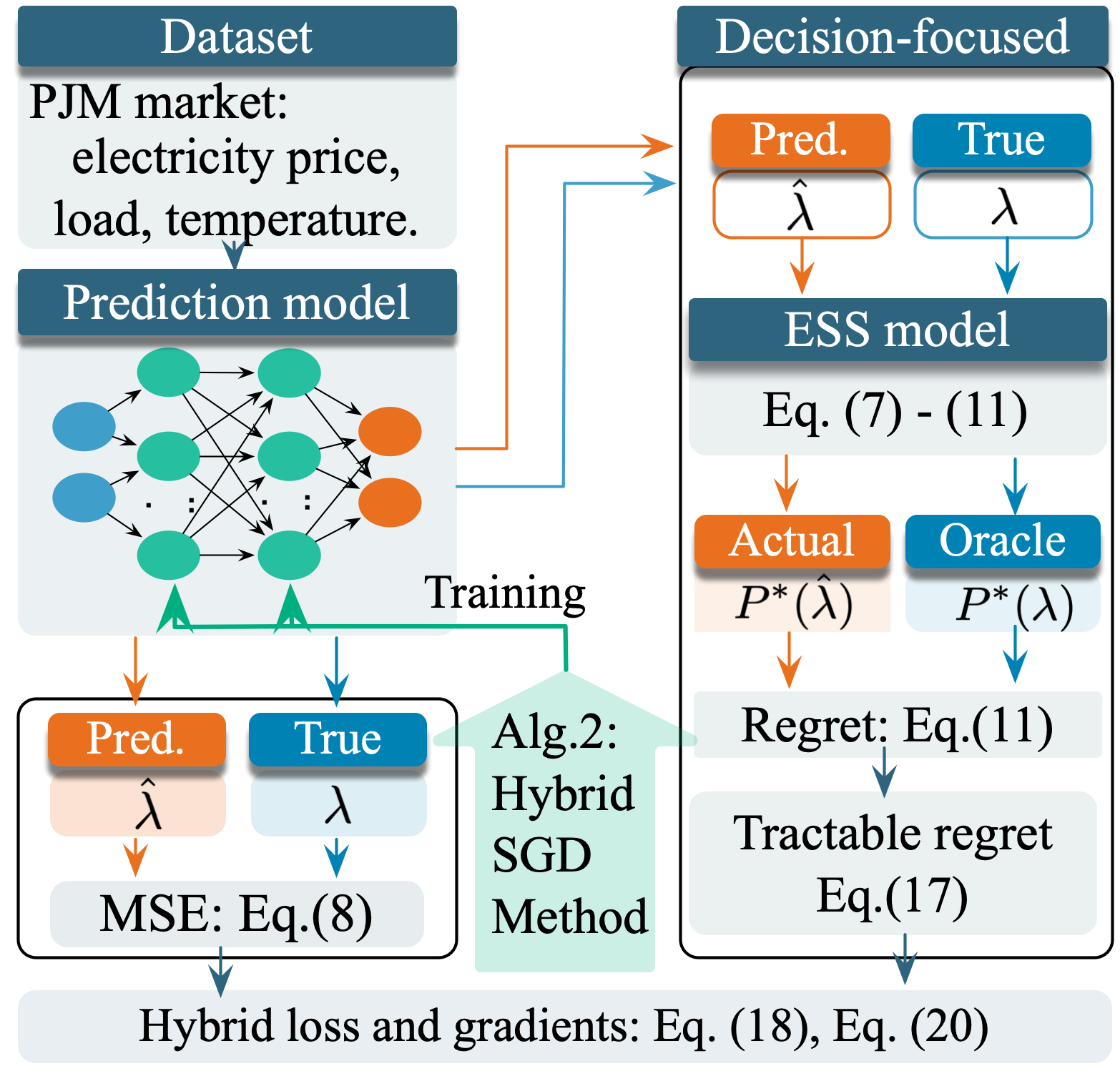}
  \caption{The decision-focused electricity price prediction approach for ESS arbitrage.}
  \label{transforming}
\end{figure}

Fig. \ref{transforming} introduces the overall decision-focused electricity price prediction approach for ESS arbitrage. As shown on the left side of Fig. \ref{transforming}, the conventional prediction-focused prediction process is based on the MSE between the predicted price and the true price. The right and bottom of Fig. \ref{transforming} present our main contributions. The decision-focused approach measures the difference between the optimal actual decision under the predicted price and the optimal oracle decision under the true price by regret and then turns it into tractable surrogate regret. Integrating MSE and tractable regret formulates the hybrid loss function, and then a hybrid SGD learning method is proposed to train prediction models.

\subsection{Regret of Decisions}
In this paper, regret describes the difference between the benefit of decisions under predicted values and that under actual values \cite{Sutton2018}, which measures the performance of online learning algorithms, \textit{e.g.}, multi-armed bandits, reinforcement learning, Thompson sampling.

\subsubsection{Regret Loss of ESS Decisions}
In the proposed electricity price prediction for ESS, we measure the difference between optimal actual decisions under the predicted price and optimal oracle decisions under the true price in Definition 1. Low regret loss means asymptotically optimal oracle decisions.
\begin{definition}[Regret of ESS decisions]
The regret of ESS decisions is defined as the gap between the benefit under the optimal actual decisions and that under the optimal oracle decisions:  
\end{definition}
\begin{equation}
  \label{gradientsofregret}
  \begin{aligned}
    regret(\hat{\lambda}, \lambda) = & \lambda^T P^*(\lambda) - \lambda^T P^*(\hat{\lambda}) \\
  \end{aligned}
\end{equation}
where $ P^*(\lambda)$, $ P^*(\hat{\lambda})$ denote the optimal oracle decisions under true price and the optimal actual decisions under the predicted price.

\subsubsection{Discussion}
\begin{figure}[ht]
  \centering
  \includegraphics[scale=0.54]{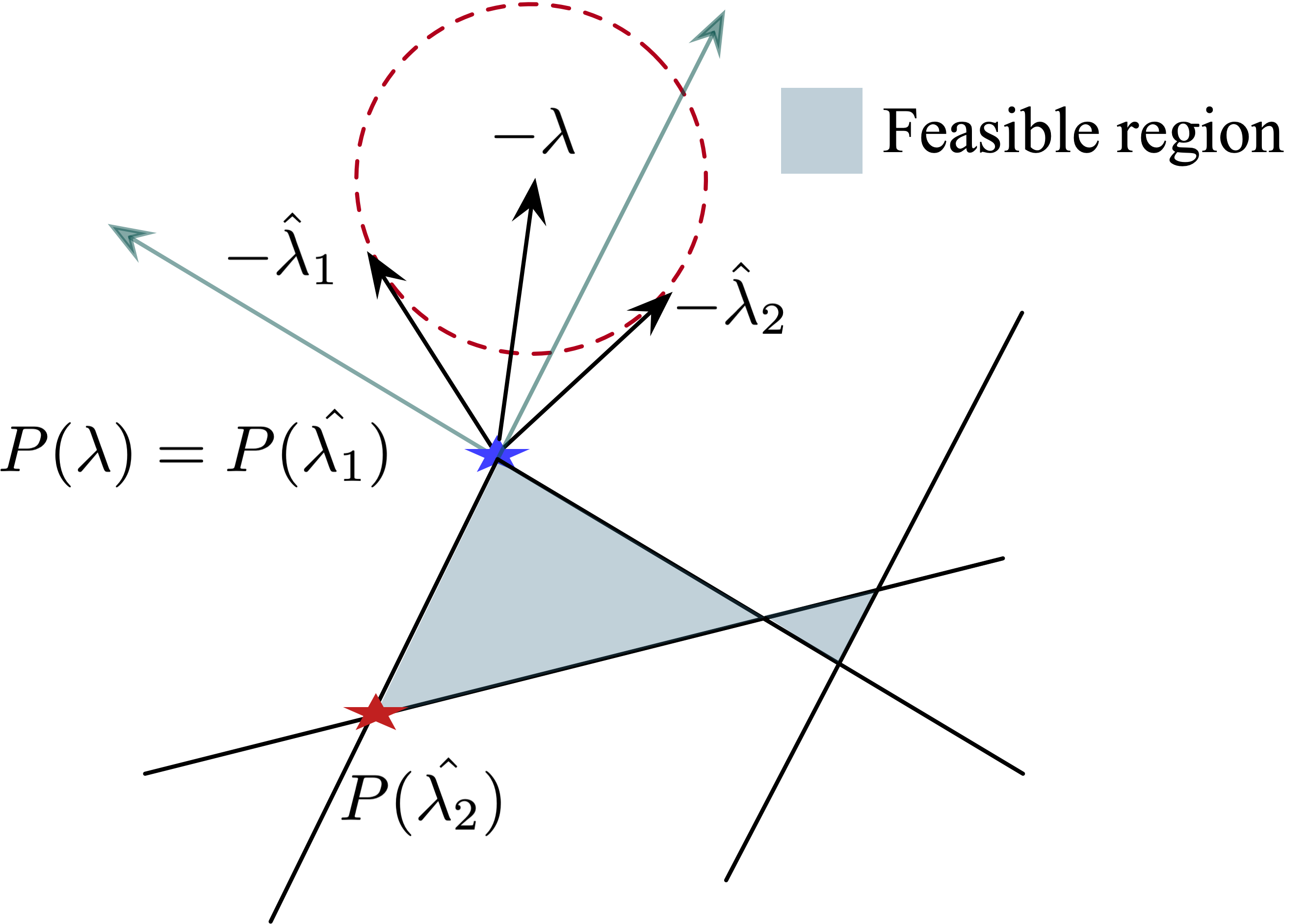}
  \caption{Geometric illustration of different predicted electricity price with same predicted error but different regrets.}
  \label{Feasible}
\end{figure}

As a typical MILP model, the ESS arbitrage model's feasible region is a collection of polyhedrons. So its optimal decision probably lies in the extreme point of one polyhedron. Taking two-dimension polyhedrons as an example, Fig. \ref{Feasible} shows that two different predicted electricity prices with the same predicted errors lead to different decision errors and corresponding different regrets. The decision $P(\lambda)$ under $\lambda$ is the same as $P(\hat{\lambda}_1)$ under $\hat{\lambda}_1$, marked by blue star, while different from $P(\hat{\lambda}_2)$ under $\hat{\lambda}_2$, marked by red star. So the prediction error is not equivalent to decision error under this scenario.

The above example illustrates that the regret of prediction focuses on the impact of prediction errors on decisions. Prediction models should consider regret in learning their parameters, but the gradients of regret to predicted price are hard to calculate directly.

\subsection{Tractable Transformation of Regret}

Regret is discontinuous and non-tractable with respect to predicted price $\hat{\lambda}$. Based on Ref. \cite{Elmachtoub2020}, we formulate the tractable regret loss function $L^{regret}$, denoted by surrogate regret.

Firstly, we derive the equivalence of regret definition in Equation (\ref{deductionofregret0}).
{\rv
\begin{equation}
  \label{deductionofregret0}
  \small
  \begin{aligned}
    regret(\hat{\lambda}, \lambda) &= \lambda^T P^*(\lambda) - \alpha \hat{\lambda}^T P^*(\hat{\lambda}) + [\alpha \hat{\lambda}^T P^*(\hat{\lambda}) - \lambda^T P^*(\hat{\lambda})]\\
    &= [\alpha \hat{\lambda}^T P^*(\hat{\lambda}) - \lambda^T P^*(\hat{\lambda})] + c^*(\lambda) - \alpha c^*(\hat{\lambda})\\
  \end{aligned}
\end{equation}
}
where $c^*(\hat{\lambda})$ and $c^*(\lambda)$ denote the optimal benefit under the predicted price $\hat{\lambda}$ and that under the true price $\lambda$ respectively.

Relaxing $P^*(\hat{\lambda})$ by the optimal decision under $\alpha \hat{\lambda} - \lambda$ will result in an upper bound of Equation (\ref{deductionofregret0}). This is satisfied for all $\alpha$, and then we derive Equation (\ref{deductionofregret1}).
{\rv
\begin{equation}
  \label{deductionofregret1}
  \small
  \begin{aligned}
    regret(\hat{\lambda}, \lambda) \leq \inf_{\alpha} \Big\{ 
      \max_{P \in \Phi} \{ 
        \alpha \hat{\lambda}^T P - \lambda^T P \} 
        - \alpha c^*(\hat{\lambda}) \Big\} 
        + c^*(\lambda) \\
  \end{aligned}
\end{equation}
}
$q(\alpha):=\max_{P \in \Phi} \{\alpha \hat{\lambda}P - \lambda P \} - \alpha c^*(\hat{\lambda})$ is actually a decreasing function of $\alpha$. According to the definition of $q(\alpha)$, a sub-gradient $g(\alpha)$ of $q(\alpha)$ is given by $g(\alpha) := \hat{\lambda} P^*(\alpha \hat{\lambda} - \lambda) - c^*(\hat{\lambda})$. {\rv $P^*(\alpha \hat{\lambda} - \lambda)$ is the optimal decision under $\alpha \hat{\lambda} - \lambda$, so its benefit $\hat{\lambda}P^*(\alpha \hat{\lambda} - \lambda)$ is less than $\hat{\lambda}P^*(\hat{\lambda})$, \textit{i.e.}, $c^*(\lambda)$.} $g(\alpha)$ is less than 0, and hence {\rv $q(\alpha)$} is a monotone decreasing function. {\rv And the $\inf_{\alpha}\{\cdot\}$ of Equation \eqref{deductionofregret1} can be replaced by $\lim_{\alpha \to \infty} \{\cdot\}$, according to the monotone converge theorem \cite{Linero-Bas2021}.} As $\alpha$ is getting larger, the term of $\lambda P$ tends to be negligible and the optimal $P$ of inner maximization problem tends to be $P^*(\hat{\lambda})$, which recovers Equation (\ref{deductionofregret0}). So the minimum upper bound of regret is established.
\begin{equation}
  \label{lossofregret1}
  \begin{aligned}
    regret^{ub}(\hat{\lambda}, \lambda) = & \lim_{\alpha \to \infty } \Big \{ \max_{P \in \Phi} \{\alpha \hat{\lambda}^T P - \lambda^T P \} - \alpha c^*(\hat{\lambda}) \Big\} \\
    & + c^*(\lambda)
  \end{aligned}
\end{equation}

When combined with previous prediction model formulation (\ref{Prediction}), the minimization of prediction models' regret can be further extended in Equation (\ref{SPO+deduction}).
\begin{figure*}[!t]
  \begin{subequations}
    \label{SPO+deduction}
    \begin{align}
      &\min_{\theta} \frac{1}{n} \sum_{i=1}^n \lim_{\alpha_i \to \infty } \Big\{ \max_{P \in \Phi} \{\lambda^T P - \alpha_i f(x_i;\theta)^T P \} - \alpha_i c^*(\hat{\lambda}) \Big\}  + c^*(\lambda) \nonumber \\
      = & \min_{\theta} \frac{1}{n} \sum_{i=1}^n \lim_{\alpha_i \to \infty } \Big\{ \max_{P \in \Phi} \{\lambda^T P - \alpha_i f(x_i;\theta)^T P \} - \alpha_i f(x_i;\theta)^T P^*(\alpha f(x_i;\theta))  + c^*(\lambda) \Big\} \\
      = & \min_{\theta}  \lim_{\alpha \to \infty } \frac{1}{n} \sum_{i=1}^n \Big\{ \max_{P \in \Phi} \{\lambda^T P - \alpha f(x_i;\theta)^T P \} - \alpha f(x_i)^T P^*(\alpha f(x_i;\theta))  + c^*(\lambda) \Big\} \\
      \leq & \min_{\theta} \frac{1}{n} \sum_{i=1}^n   \max_{P \in \Phi} \Big\{\lambda^T P - 2 f(x_i;\theta)^T P \Big\} - 2 f(x_i;\theta)^T P^*(2 f(x_i;\theta)) + c^*(\lambda)\\
      \leq & \min_{\theta} \frac{1}{n} \sum_{i=1}^n   \max_{P \in \Phi} \Big\{\lambda^T P - 2 f(x_i;\theta)^T P \Big\} - 2 f(x_i;\theta)^T P^*(\lambda) + c^*(\lambda)
    \end{align}
  \end{subequations}
\end{figure*}
The first equality (\ref{SPO+deduction}a) of above derives from the fact $P^*(\alpha_i f(x_i)) = P^*(f(x_i))$ for any $\alpha_i \geq 0$, which is proved in (\ref{eq: proof2}). The second equality (\ref{SPO+deduction}b) derives from the intuition that as all $\alpha_i$ tends to be $\infty$, so they tend to be the same and can be replaced by a single variable $\alpha$. The first inequality (\ref{SPO+deduction}c) derives from setting $\alpha$ as 2 to get an estimate of upper bound in particular. And the second inequality (\ref{SPO+deduction}d) relaxes optimal value of $P^*(2 f(x_i))$ under $c(\hat{\lambda})$  by a feasible decision value $P^*(\lambda)$.

\begin{eqnarray}
  \begin{aligned}
    P^*(\alpha_i f(x_i)) &:= \arg \max_{P \in \Phi} \alpha_i f(x_i)^T P \\ 
    & = \alpha_i \arg \max_{P \in \Phi}  f(x_i)^T P \\ 
    & = \alpha_i P^*(f(x_i))
  \end{aligned}
  \label{eq: proof2}
\end{eqnarray}

\begin{definition}[Surrogate regret loss of ESS decision]
Given the predicted price $\hat{\lambda}$ and true price $\lambda$, the surrogate regret loss of ESS decision can be defined as:
\end{definition}

\begin{equation}
  \label{lossofregret2}
  \begin{aligned}
    L^{regret}(\hat{\lambda}, \lambda) &= \max_{P\in \Phi_{ESS}} \{ \lambda^T P - 2 \hat{\lambda}^T P \} - 2 \hat{\lambda} P^*(\lambda) + c^*(\lambda)  \\
    &= (\lambda - 2 \hat{\lambda})^T P^*(\lambda - 2 \hat{\lambda}) - 2 \hat{\lambda} P^*(\lambda) + c^*(\lambda) \\
  \end{aligned}
\end{equation}

\begin{remark}[Properties of surrogate regret loss]
  Given predicted price and true price, proposed surrogate regret loss holds the following properties:

  1. $regret(\hat{\lambda}, \lambda) \leq L^{regret}(\hat{\lambda}, \lambda)$;

  2. $L^{regret}(\hat{\lambda}, \lambda)$ is a convex function of predicted electricity prices $\hat{\lambda}$.
\end{remark}

\begin{lemma}[Gradients of the surrogate regret to predicted prices]
Based on the surrogate regret definition, the gradient of regret to predicted price can be derived as:
\end{lemma}
\begin{equation}
  \label{SPO+grad}
  \begin{aligned}
    \frac{\partial L^{regret}(\hat{\lambda}, \lambda)}{\partial \hat{\lambda}} &= - 2 (P^*(\lambda) + P^*( \lambda - 2\hat{\lambda}))\\ 
  \end{aligned}
\end{equation}

\begin{equation*}
  \label{SPO+proof}
  \begin{aligned}
    \text{Left side} &= \frac{\partial \max_{P\in \Phi} \{ \lambda^T P - 2 \hat{\lambda}^T P \} - 2 \hat{\lambda} P^*(\lambda) + c^*(\lambda)}{\partial \hat{\lambda}} \\ 
    &= \frac{\partial \max_{P\in \Phi} \{ \lambda^T P - 2 \hat{\lambda}^T P \}}{\partial \hat{\lambda}} - 2 P^*(\lambda) \\
    &= -2 P^*(\lambda - 2 \hat{\lambda}) - 2 P^*(\lambda)
  \end{aligned}
\end{equation*}

\subsection{Hybrid SGD Learning Method for ESS Arbitrage}

\begin{figure}[ht]
  \centering
  \includegraphics[scale=0.8]{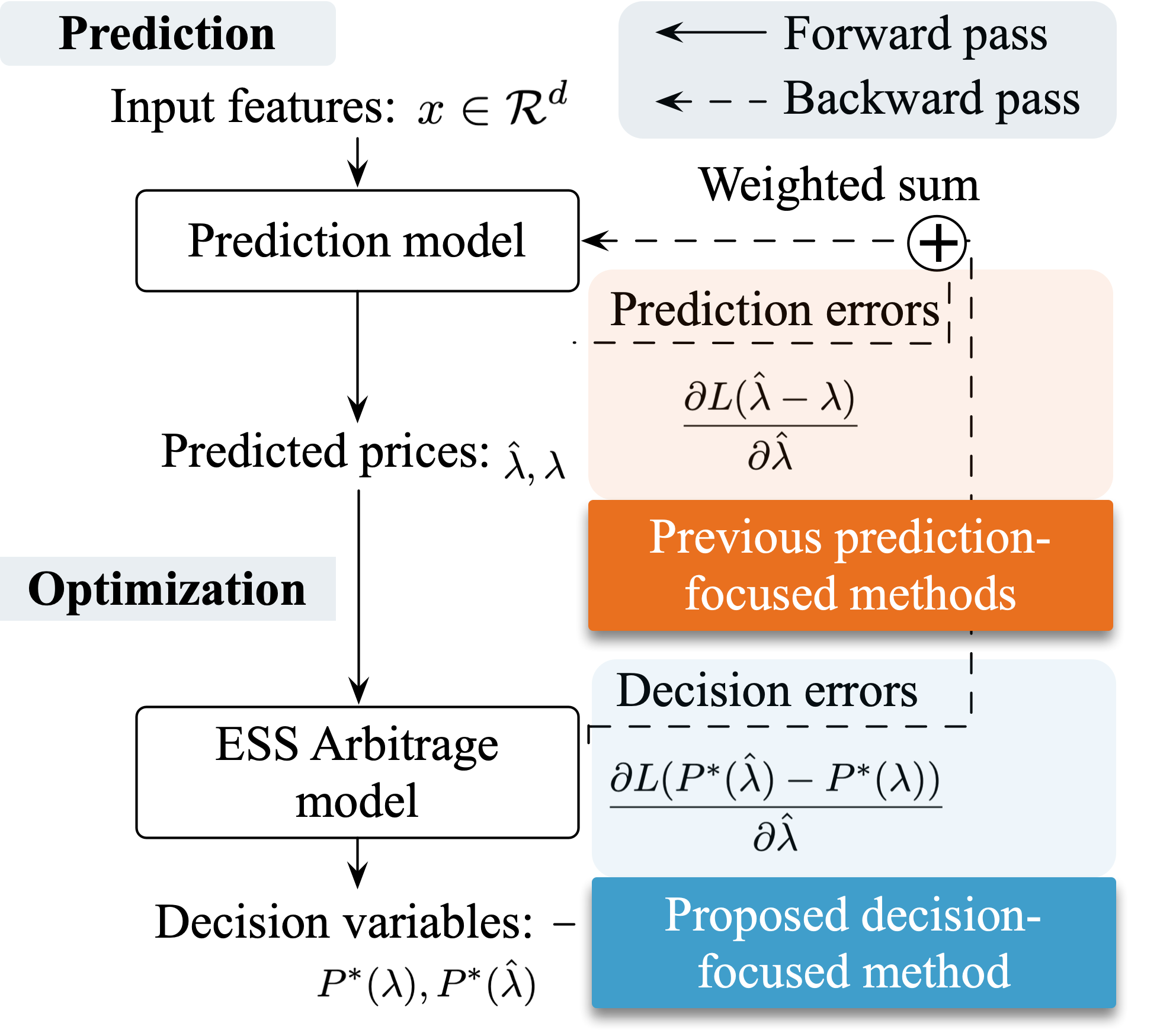}
  \caption{The scheme of the hybrid SGD learning method for ESS arbitrage.}
  \label{fig: method}
\end{figure}

MSE and surrogate regret focus on the prediction errors and decision errors individually, which view prediction mismatches from the perspectives of the prediction model and the ESS model. So this paper designs a hybrid loss function and proposes a hybrid stochastic gradient descent (SGD) learning method for training prediction model based on the hybrid loss function (\ref{SPO+comb+loss}) and its derivatives (\ref{SPO+comb}). Fig. \ref{fig: method} presents the scheme of the hybrid SGD learning method.

\begin{definition}[Hybrid loss function]
  The hybrid loss can be defined as the weighted sum of MSE and surrogate regret to capture decision and prediction errors.
\end{definition}

\begin{equation}
  \label{SPO+comb+loss}
  \begin{aligned}
    L^{comb}(\hat{\lambda}, \lambda) &= L^{regret}(\hat{\lambda}, \lambda) +\epsilon L^{MSE}(\hat{\lambda}, \lambda) \\
    &= (\lambda - 2 \hat{\lambda})^T P^*(\lambda - 2 \hat{\lambda}) - 2 \hat{\lambda} P^*(\lambda) + c^*(\lambda) \\
    &\quad + \epsilon \frac{1}{T} \frac{||\hat{\lambda} - \lambda ||^2_2}{2}
  \end{aligned}
\end{equation}
where $\epsilon$ is a weighted coefficient on MSE, which implies the emphasis extent of {\rv minimizing} prediction errors.

Then we derive the hybrid gradients of combining regret and MSE to predicted price as:
{\rv
\begin{equation}
  \label{SPO+comb}
  \begin{aligned}
    \frac{\partial L^{comb}(\hat{\lambda}, \lambda)}{\partial \hat{\lambda}} & = \frac{\partial L^{regret}(\hat{\lambda}, \lambda)}{\partial \hat{\lambda}}+\epsilon \frac{\partial L^{MSE}(\hat{\lambda}, \lambda)}{\partial \hat{\lambda}} \\
    = - 2 (P^*&(\lambda) + P^*( \lambda - 2\hat{\lambda})) + \frac{1}{T} \epsilon \sum_{t=1}^T (\hat{\lambda}_t - \lambda_t) 
  \end{aligned}
\end{equation}
}

The gradients of the hybrid loss function can be calculated explicitly by (\ref{SPO+comb}). {\rv To improve the calculation efficiency, the gradients of MSE are calculated implicitly by the automatic differentiation algorithm \cite{Paszke2017} from the Autograd tool \cite{Paszke2019} in Pytorch. Autograd is a reverse automatic differentiation system for calculating derivatives, which is the core of Pytorch \cite{Paszke2019}. It records a graph of all the operations performed on a gradient-enabled tensor and creates an acyclic graph called the dynamic computational graph. Gradients are computed by tracing the graph from the root to the leaf and multiplying each gradient through the chain rule.

The automatic differentiation algorithm of Autograd features high calculation efficiency due to its core logic in C++ \cite{Paszke2017}, which can calculate the MSE gradient implicitly in a more efficient way than the explicit calculation by Equation \eqref{SPO+comb}.} In contrast, the gradients of surrogate regret are calculated explicitly by solving the MILP problem in (\ref{SPO+grad}). 

To tackle these two distinct ways of calculating gradients, the prediction model training process under hybrid gradients is divided into three steps: 1) calculate the weighted MSE loss of predicted price, back-propagate the MSE loss to the parameters, and retain the gradients for later parameters gradients updating; 2) calculate the gradients of surrogate regret based on Equation (\ref{SPO+grad}), feed the gradients to the predicted price, and back-propagate the gradients to the same parameters of the prediction model; 3) update the prediction model parameters based on accumulated gradients. The hybrid gradients updating is achieved by twice back-propagating and only once updating.

Based on the above derivation and discussion, this paper proposes the hybrid SGD learning method for ESS arbitrage to achieve the prediction and decision accuracy in Algorithm 1. We note that the proposed SGD learning method can apply to simple linear prediction models and complex deep learning models.

\begin{algorithm}[!t]
  \label{CombSGD}
  \caption{Hybrid SGD learning method for ESS arbitrage}
  {\small{
    \begin{algorithmic}[1]
      \STATE \textbf{Input:} Raw dataset, prediction model; hyperparameter: batch size $N$, learning rate $\alpha$,{\rv training epochs $M$}; 
      \STATE \textbf{Data processing:} Data pre-processing, featuring engineering, and train-validation-test dataset dividing; 
      \STATE \textbf{Model initializing:}
      \STATE \textbf{Training:}
      \FOR{{\rv $m$ = 1, ... $M$}}
      \STATE Initialize the gradients of parameters: $\frac{\partial L^{Comb}(\hat{\lambda}, \lambda)}{\partial \theta} = 0$
      \STATE Sampling $N$ data point; 
      \FOR{$i$ = 1, ... $N$}
      \STATE Predict price: $\hat{\lambda_i} = f(x_i)$;
      \STATE \textbf{Calculate} the weighted gradients of MSE $L_i^{MSE}(\hat{\lambda}_i, \lambda_i)$ based on the Autograd tools;
      \STATE \textbf{1st back-propagate} the weighted gradients of MSE to the parameters of prediction model; $$\frac{\partial L^{Comb}(\hat{\lambda}, \lambda)}{\partial \hat{\theta}} = \frac{\partial L^{Comb}(\hat{\lambda}, \lambda)}{\partial \hat{\theta}} + \epsilon \frac{\partial L_i^{MSE}(\hat{\lambda}_i, \lambda_i)}{\partial \hat{\theta}}$$
      \STATE \textbf{Calculate} gradients of surrogate regret: $$\frac{\partial L^{regret}(\hat{\lambda}_i, \lambda_i)}{\partial \hat{\lambda}_i} =  - 2 (P^*(\lambda_i) + P^*( \lambda_i - 2\hat{\lambda}_i))$$
      \STATE \textbf{2nd back-propagate} the gradients of surrogate regret to the parameters of prediction model;$$\frac{\partial L^{Comb}(\hat{\lambda}, \lambda)}{\partial \theta} = \frac{\partial L^{Comb}(\hat{\lambda}, \lambda)}{\partial \theta} + \epsilon \frac{\partial L^{regret}(\hat{\lambda}_i, \lambda_i)}{\partial \hat{\lambda}_i} \frac{\partial \hat{\lambda}_i}{\partial \theta}$$
      \ENDFOR
      \STATE \textbf{Update} model parameters in batches: $$\theta = \theta - \alpha \frac{\partial L^{Comb}(\hat{\lambda}, \lambda)}{\partial \theta}$$
      \STATE \textbf{Evaluate} prediction model in the validation dataset;
      \ENDFOR 
      \STATE \textbf{Output:} Final prediction model.
    \end{algorithmic}
  }}
\end{algorithm}

\section{Case Study}
This paper utilizes six-year hourly actual electricity price, related temperature, and load datasets from the Pennsylvania-New Jersey-Maryland (PJM) interconnection \cite{PJM} to verify the effectiveness of the proposed decision-focused approach. All the proposed models and methods are implemented by python equipped with Pytorch package \cite{Paszke2019} for prediction models, with Cvxpy package \cite{Diamond2016} for optimization models. The experiment computer is a MacBook Pro laptop with RAM 16 GB, CPU Intel Core I7 (2.6GHz).

This section firstly pre-processes the raw electricity-related data to construct the prediction dataset, then extracts the input-output vectors, trains different day-ahead prediction models, and evaluates various models with different loss functions by multiple metrics. 

\subsection{Datasets and Models}

\subsubsection{Data Preparing}
We construct the input features from three aspects in Table I: i) the past day information: the hourly load, temperature, and temperature square in the past day; ii) the prediction day information: the prediction temperature and its square in the prediction day; {\slw iii)} calendar effects: the indicators of weekday, holiday, and day of the year.

Then all the input features are standardized to formulate the final feature vectors based on the mean and standard variance.

As the electricity price fluctuates a lot, we set the output of the prediction model as the log of electricity price, which assumes electricity price follows log-normal distribution \cite{Donti2017}. After feature engineering, the 20\% of the pre-processed datasets are split randomly into the test set for model evaluation; the 20\% of the rest are further split randomly into the validation set for training early stop; the final rest is the training set for learning the parameters of the model. 

\subsubsection{Models Initializing}

\begin{figure}[ht]
  \centering
  \includegraphics[scale=0.7]{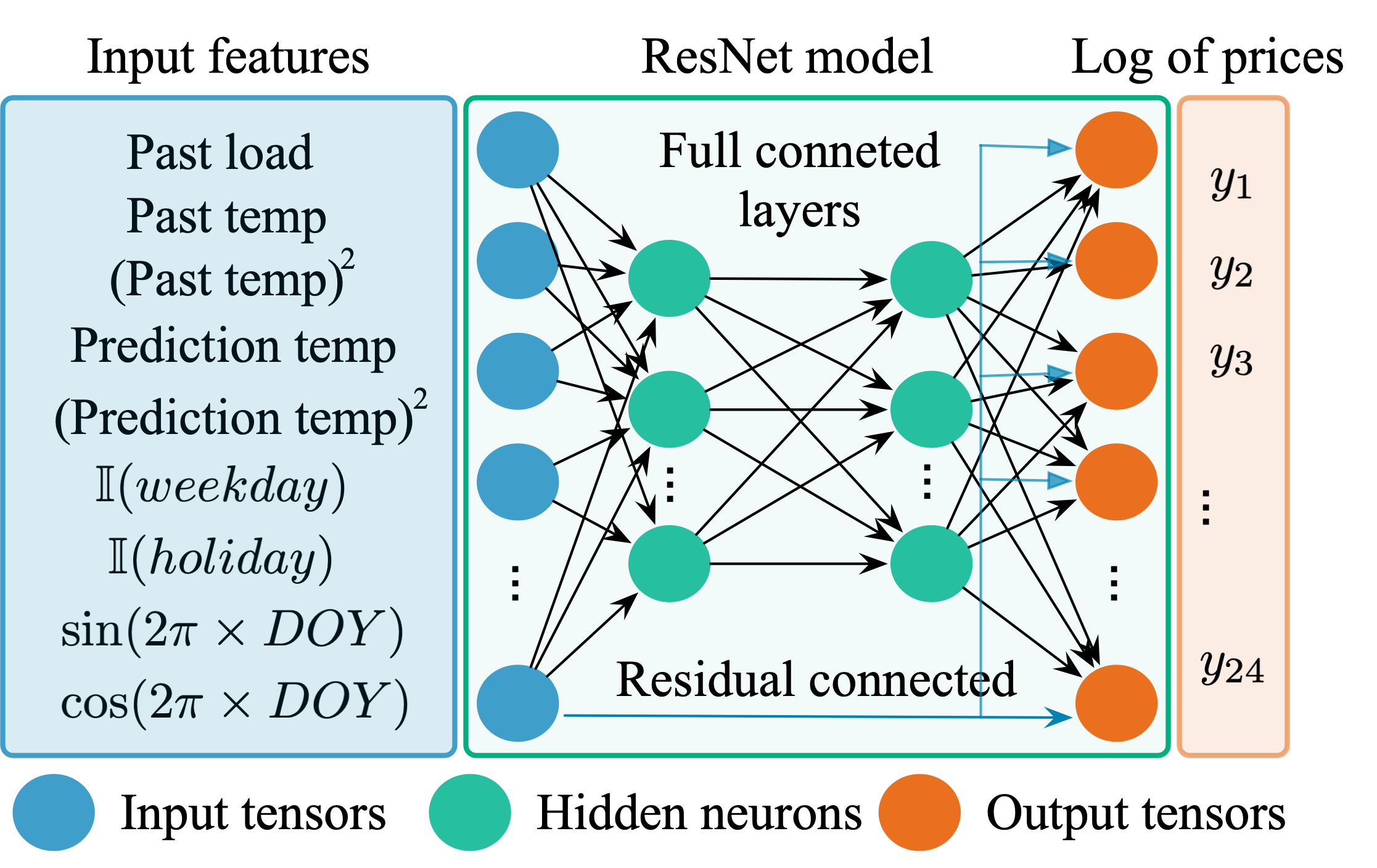}
  \caption{The structure of ResNet model.}
  \label{structure}
\end{figure}

Fig. \ref{structure} visualizes the whole structure of the prediction process, whose key lies in the design of the ResNet model, which maps constructed input features to output log of price. As shown in Fig. \ref{structure}, the ResNet model is composed of full connected (FC) layers and residual connected (RC) layers. The parameters of FC layers are initialized randomly, and whereas the parameters of RC layers are initialized by the least square method $(X^T X)^{-1} X^T Y$.

For the ESS arbitrage model, we normalize the energy capacity into 1 MWh. The charging/discharging power depends on the daily depths of charging/discharging, where the depths refer to the ratios of maximum charging/discharging power (MW) to energy capacity (MWh). So the unit of regret in our case is \$/MWh.

\subsubsection{Hyperparameter Setting}
The hyperparameter setting includes the ResNet model and ESS model hyperparameters in Table I. 

\begin{table}[ht]
  \renewcommand{\arraystretch}{1.3}
  \centering
  \caption{{\rv The hyperparameters of ResNet prediction and ESS arbitrage models.}}
  \label{hyperparameters}
  \begin{tabular}{cc|cc}
    \hline
    \bfseries{ResNet model} & & \bfseries{ESS arbitrage model} & \\
    \hline
     Hyper-parameters & Values & Hyper-parameters & Values \\
    \hline
     Optimizer & Adam & $E_{min}$ / $E_{max}$ & 0.2 / 0.95 \\
     Learning rate & 1e-6 & Depth of charging & 0.5  \\
     Hidden layers & [50, 50] & Depth of discharging & 0.5 \\
     Batch size& 100 & $\eta_{ch}$ & 0.90 \\
     Dropout & 0.2 & $\eta_{dis}$ & 0.92 \\
     {\rv $\epsilon$} & {\rv 25} & {\rv $\Delta_t$} & {\rv 1 hour} \\  
     {\rv Training epochs} & {\rv 50} & & \\
    \hline
  \end{tabular}
\end{table}

\subsection{Performance of the Decision-focused Prediction Model}

\subsubsection{Training Process}

\begin{figure}[ht]
  \centering
  \subfigure[The MSE changing process]{\includegraphics[scale=1]{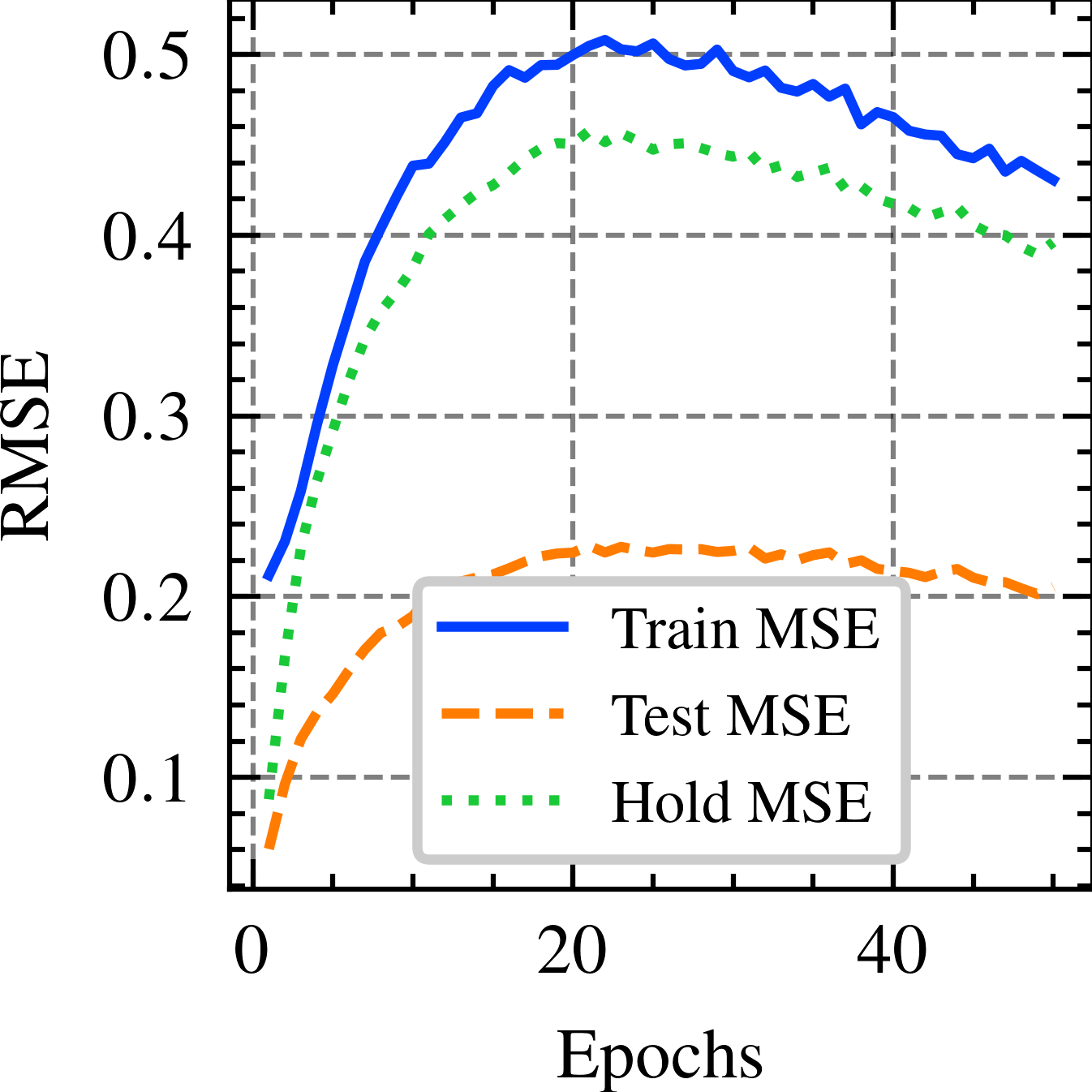}}
  \qquad
  \subfigure[The regret changing process]{\includegraphics[scale=1]{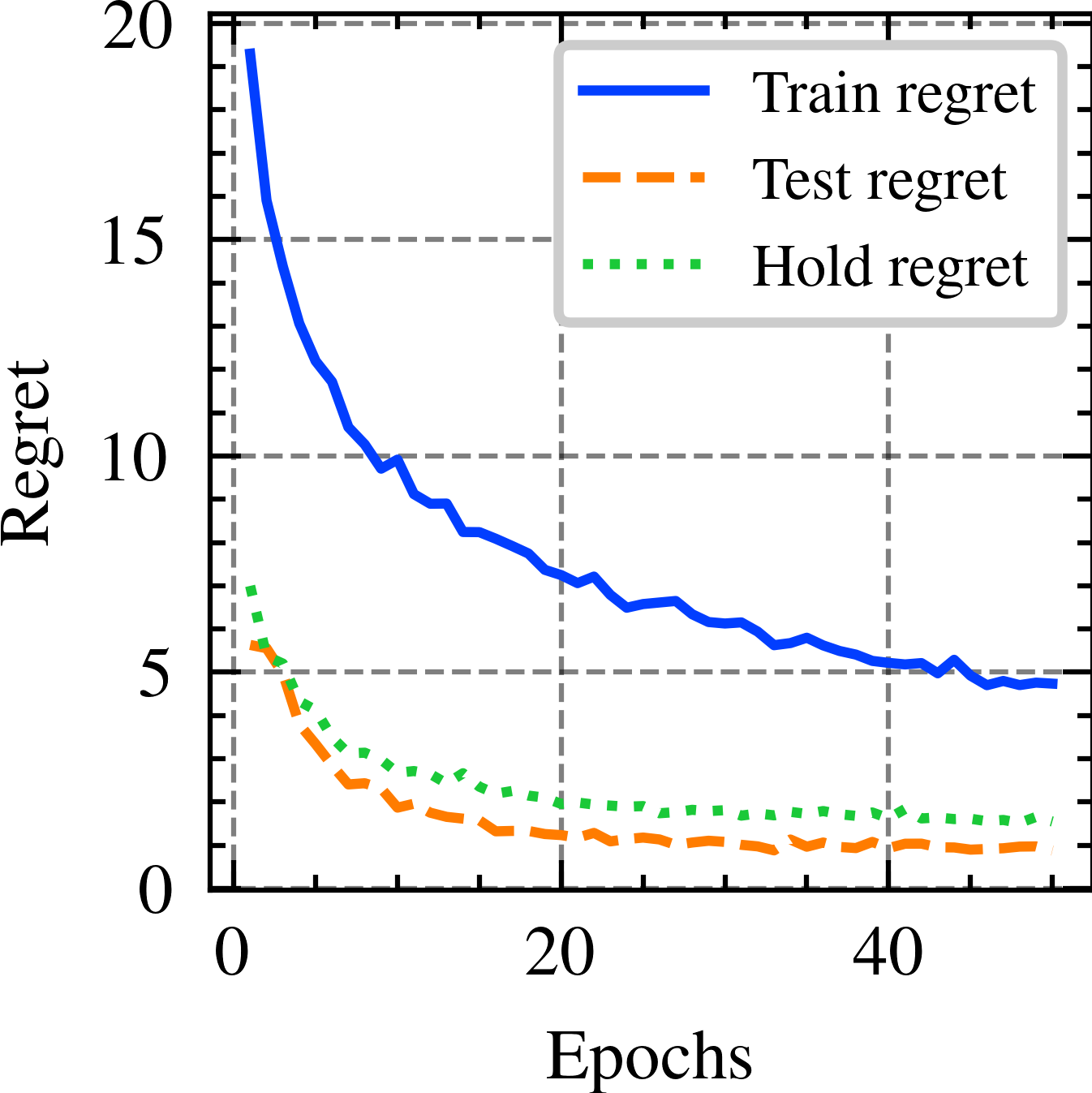}} 
  \caption{The errors changing in the training process.}
  \label{Training_process}
\end{figure}

Fig. \ref{Training_process} visualizes MSE and regret changing in training the decision-focused ResNet prediction model. As training epochs increase, the MSE of three split datasets increase firstly and then decrease at around the 23rd epoch in Fig. \ref{Training_process} (a). In contrast, the regrets of three split datasets decrease shapely at first and then flatten at around 45th epoch in Fig. \ref{Training_process} (b). 

The different loss changing curves of MSE and regret show that predicted price features low MSE but high regret under initialized parameters of ResNet. This phenomenon implies that MSE is not consistent with regret. The mere consideration of MSE loss function in training prediction models is not comprehensive.

\subsubsection{Models' Comparison}
\begin{figure}[ht]
  \centering
  \includegraphics[scale=1]{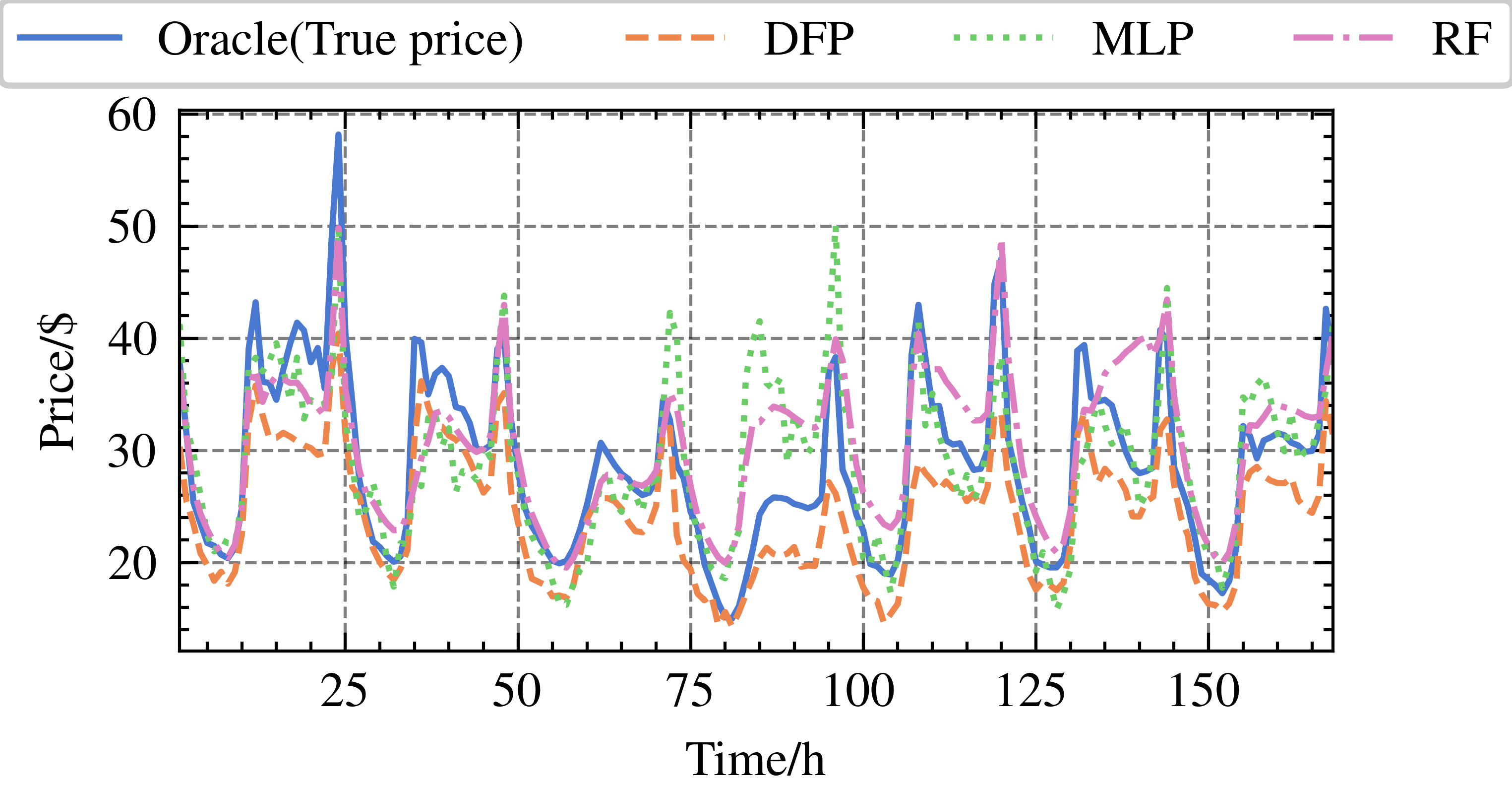}
  \caption{\rv Predicted price under different methods.}
  \label{fig: prediction compare}
\end{figure}

To verify the effectiveness of the proposed approach, we compare the proposed decision-focused prediction (DFP) model (considering hybrid loss) with the MSE-based model (only considering MSE), multi-layer perceptron (MLP), and random forest (RF) model. The measuring metrics of prediction models' performance in test set mainly have two components: prediction accuracy metrics, \textit{i.e.}, root-mean-square-error (RMSE), mean-absolute-percentage-error (MAPE) and decision accuracy metrics, \textit{i.e.}, regret and benefits of ESS arbitrage. We note that benefits refer to arbitrage profits from price fluctuation under different prediction models. {\rv The affinity propagation algorithm \cite{Frey2007} is utilized to acquire the typical week electricity price and visualize changing trends of true prices and predicted prices in Fig. \ref{fig: prediction compare}, where the oracle (true price) is the benchmark for comparison.} As shown in the highlighted area of Fig. \ref{fig: prediction compare}, the proposed DFP method can predict the changing trend of oracle price more accurately than MLP and RF models, which is the key of accurate decision. {\rv We note that the solid blue line with the ``oracle'' label refers to the true price, which is the benchmark to evaluate the prediction performance.}

Then we analyze the results of different prediction models in Table II. The proposed DFP model achieves lower decision regret than other methods, though it endures high MSE and MAPE metrics than the MLP and the random forest methods. DFP model is apparently superior to the MSE-based model by 90.91\% less regret and 46.8\% higher benefits.{\rv We note that the benefits of different methods in this paper are actually the average daily benefits in the testing set, which is the mean benefits of the decisions from different models in all the testing days. The average arbitrage benefits under the true price without prediction error are \$30.712 for a 500-kWh ESS, and the benefits under different models in the case study are scalable to ESSs with different capacities.} Compared with MLP methods, DFP reduces 40.3\% regret and further increases \$1.72 (about 6.11\%) daily benefits per MWh on average; compared with random forest model, DFP reduces 33.56\% regret and further increases \$0.090 (about 3.2 \%) daily benefits per MWh on average.
\begin{table}[ht]
  \renewcommand{\arraystretch}{1.3}
  \centering
  \label{models_comparison}
  \caption{Comparison of different prediction models.}
  \begin{tabular}{ccccc}
    \hline
    Prediction models &  RMSE & MAPE & Regret & Benefits(\$) \\
    \hline
    DFP model & 0.294 & 0.0779 & 0.952 & 29.86 \\
    MSE-based model & 0.320 & 0.0780 & 10.470 & 20.33\\
    MLP model & 0.0287 & 0.0412 & 1.595 & 28.14\\
    RL model & 0.0267 & 0.0415 & 1.433 & 28.94\\
    \hline
  \end{tabular}
  \vspace{1ex}

  {\raggedright Note: the hidden layer of MLP is set as 700; the depth of the random forest method is set as 30.\par}
\end{table}

\begin{figure}[ht]
  \centering
  \includegraphics[scale=1]{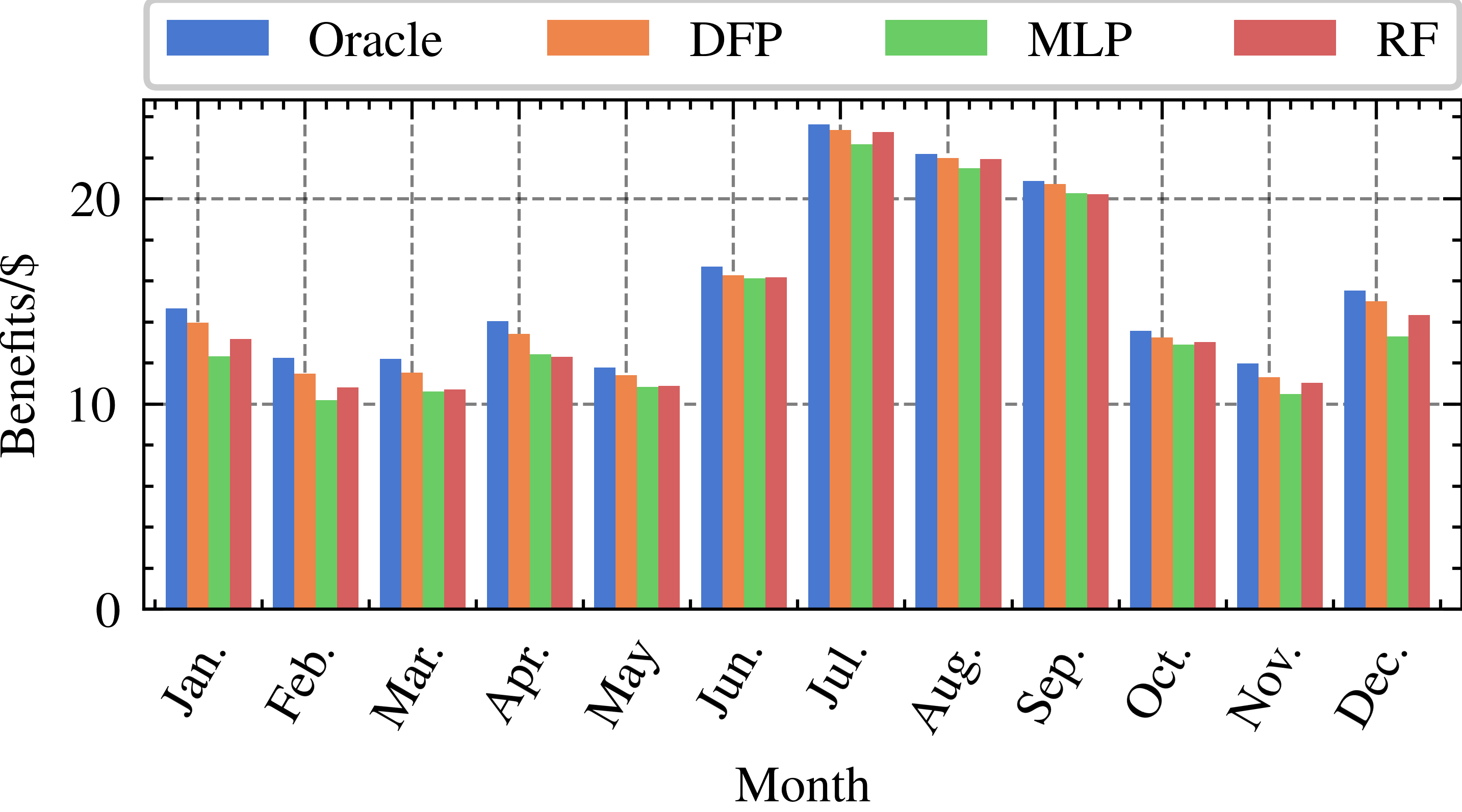}
  \caption{Benefits of 500 kWh battery under different prediction methods.}
  \label{fig: benefits compare}
\end{figure}

{\rv Taking the 500 kWh Li-ion as an example, the detailed hyperparameters of in the 500 kWh ESS arbitrage model \eqref{ESbigM}-\eqref{Contrs3} can be derived with $E_{min}$, $E_{max}$ as 10 kWh, 475 kWh, with $P^{max}_{ch}$ as 250 kW, and with $P^{max}_{dis}$ as 250 kW according to the second column of Table \ref{hyperparameters}. We note the depth of charging/discharging reflects the fraction of the capacity added/removed from the fully charged battery, which is utilized to calculate the $P^{max}_{ch}$/ $P^{max}_{dis}$.} We further compare the average daily ESS benefits from various prediction models in different months in Fig. \ref{fig: benefits compare}. {\rv The method that the day-ahead electricity price is accurately obtained without prediction error (this is impossible in practice) is taken as the benchmark for comparison, denoted by oracle, whose benefit is the highest due to no prediction error.} Fig. \ref{fig: benefits compare} presents the benefits from July to September are higher than other months due to the high fluctuation in these summer months. The benefit from DFP is closer to that from the oracle model, higher than the other two models. In April, the benefit from DFP acquires \$1.60 higher daily energy arbitrage than MLP and \$1.10 higher than RF. In the whole year, the benefit from DFP can increase \$87.78 compared to RF.

\subsection{Analysis of Decision-focused Approach}

This part utilizes the proposed hybrid SGD learning method to train the linear model (\ref{Linear}) and ResNet model (\ref{ResNet}) individually to unfold the impact of hybrid loss in different representational capacity models. In each prediction model, we further analyze the effect of different weight parameters $\epsilon$ in the hybrid loss for deciding its proper value. {\rv The weight parameter $\epsilon$ reflects the trade-off between the prediction error and decision error, which can be selected based on the plenty of experiments in Table III and IV.}

\subsubsection{Linear Prediction Model}

The proposed hybrid loss is utilized for training the simple linear prediction models (\ref{Linear}) with different $\epsilon$. For comparison, the surrogate regret loss means $\epsilon$ takes the value of 0, whereas MSE loss means $\epsilon$ takes a significant value to ignore the effects of surrogate regret part in the hybrid loss.

\paragraph{Performance evaluation}
After 100 times training, we evaluate the trained model by the same metrics in the previous part: RMSE, MAPE, regret, and benefits of ESS.

\begin{table}[ht]
  \renewcommand{\arraystretch}{1.3}
  \centering
  \label{linear_comparison}
  \caption{\rv Loss function comparison of linear prediction model.}
  \begin{tabular}{cccccc}
    \hline
    Loss function & $\epsilon$& RMSE & MAPE & Regret & Benefits(\$) \\
    \hline
    MSE & / & 3.395 & 1.03 & 18.780 & 12.02\\ 
    Surrogate regret& / & 3.239 & 0.985 & 16.214 & 14.59\\
    {\rv Hybrid loss} & {\rv 0.5} & {\rv 3.305} &{\rv 1.011} & {\rv 14.567} & {\rv 16.23}\\
    {\rv Hybrid loss} & {\rv 1 } & {\rv 3.352} & {\rv 1.025} & {\rv 14.901} & {\rv 15.90}\\
    Hybrid loss & 25 & 3.229 & 0.988 & 7.018 & 23.79\\
    Hybrid loss & 50 & 2.558 & 0.783 & 3.133 & 27.67 \\
    Hybrid loss & 100 & 1.586 & 0.482 & 2.154 & 28.65 \\
    Hybrid loss & 200 & 0.648 & 0.19 & 2.425 & 28.38 \\
    \hline
  \end{tabular}
\end{table}

As shown in Table III, with the increase of $\epsilon$ from 25 to 200, the RMSE and MAPE decrease correspondingly. However, the regret of ESS decreases at first and then increases when $\epsilon$ is 200. In contrast, MSE and surrogate regret loss models feature high RMSE, MAPE, regret, and low benefits. The proposed hybrid loss can help instruct the small representational capacity prediction models to predict and make decisions more accurately, which leads to high benefits. {\rv When $\epsilon$ is 1, the regret of the linear prediction model has doubled than the regret with $\epsilon$ as 25, while the RMSE and MAPE increase a little.}

\paragraph{Errors in different time intervals} 
When $\epsilon$ is 25, the RMSE and MAPE of MSE, surrogate regret, hybrid loss are similar, but hybrid loss achieves lower regret compared to the others.

\begin{figure}[ht]
  \centering
  \includegraphics[scale=1]{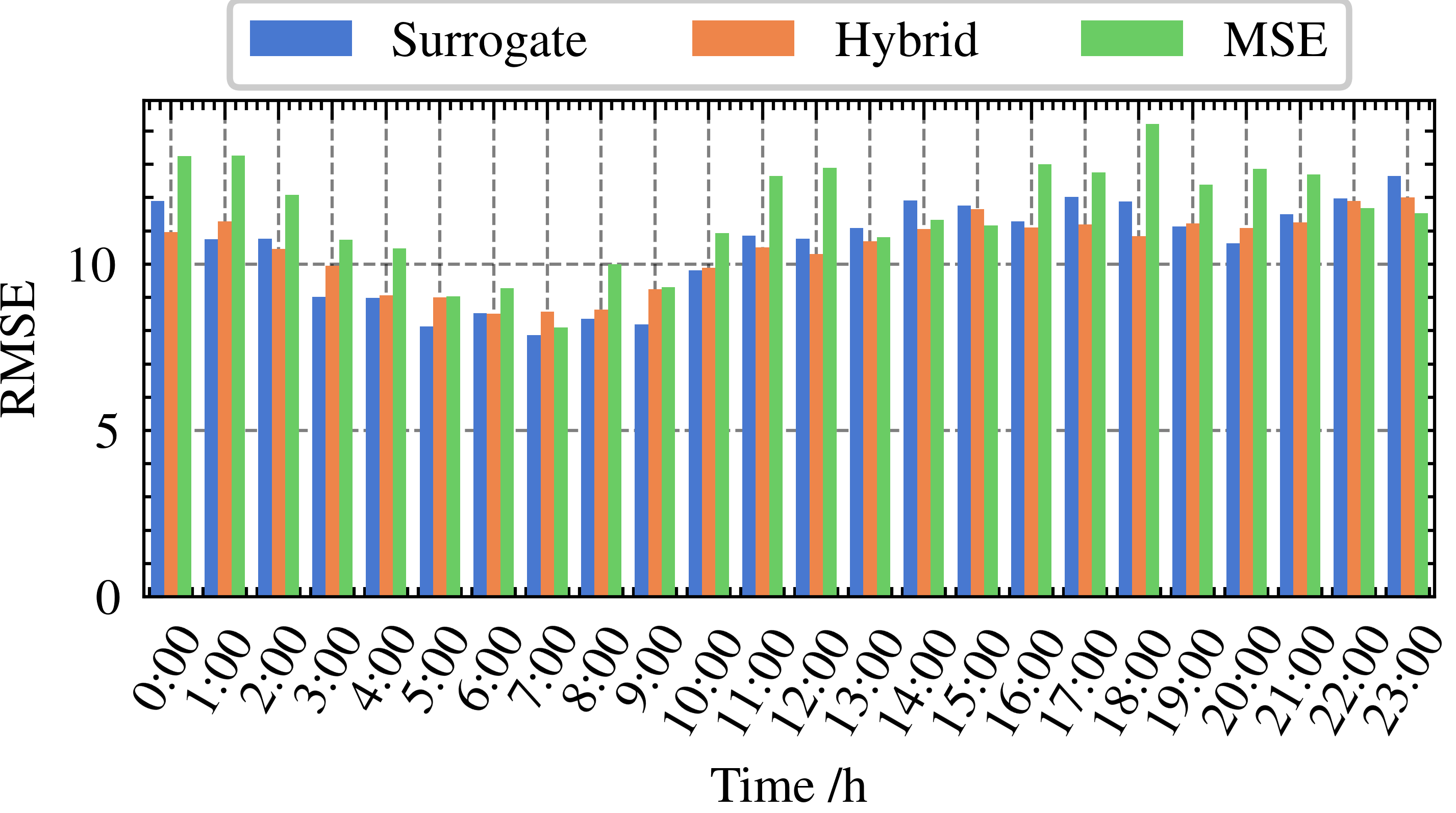}
  \caption{RMSE in different time interval of different loss function by linear model.}
  \label{interval_comparison_linear}
\end{figure}

As shown in Fig. \ref{interval_comparison_linear}, though the whole prediction errors under the above three losses are similar, the daily time distribution of prediction errors under MSE and surrogate regret embodies high variation than that under hybrid loss. The difference of error distributions verifies the effectiveness of hybrid loss for instructing the prediction model to reduce decision error.

\paragraph{Results analysis} 
The above numerical experiments provide valuable insights into the proposed hybrid loss in small representational capacity models. i) Hybrid loss can take advantage of the gradient information from prediction and decision errors to achieve more accurate prediction and decision. ii) With the increasing weight on MSE, the prediction errors are reducing alone while decision errors go down first and up after. This phenomenon can be interpreted as the hard predictable case, where electricity price is hard to predict accurately due to its uncertainty. In this case, the hybrid loss can help reduce prediction and decision errors.

\subsubsection{ResNet Prediction Model}
Then we apply the hybrid loss to train the complex ResNet models (\ref{ResNet}) with different $\epsilon$.

\paragraph{Performance evaluation}
After 50 times training, we evaluate the trained ResNet model by the same metrics previously. Table IV compares the prediction and decision performance of different loss functions. 

\begin{table}[ht]
  \renewcommand{\arraystretch}{1.3}
  \centering
  \label{ResNet_comparison}
  \caption{\rv Loss function comparison of ResNet prediction model.}
  \begin{tabular}{cccccc}
    \hline
    Loss function & $\epsilon$ & RMSE & MAPE & Regret & Benefits(\$) \\
    \hline
    MSE & / & 0.320 & 0.078 & 10.470 & 20.34 \\
    Surrogate regret & / & 0.578 & 0.094 & 0.899 & 29.91 \\
    {\rv Hybrid loss} & {\rv 0.5} & {\rv 0.427} & {\rv 0.0779} & {\rv 0.875} & {\rv 29.93} \\
    {\rv Hybrid loss} & {\rv 1} & {\rv 0.658} & {\rv 0.142} & {\rv 0.924} & {\rv 29.88} \\
    Hybrid loss & 25 & 0.294 & 0.0779 & 0.952 & 29.86 \\
    Hybrid loss & 50 & 0.199 & 0.0466 & 1.343 & 29.46 \\
    Hybrid loss & 100 & 0.153 & 0.035 & 1.938 & 28.87 \\
    Hybrid loss & 200 & 0.144 & 0.03185 & 3.588 & 27.22 \\
    \hline
  \end{tabular}
\end{table}

Similar to linear models, with the increase of $\epsilon$ from 25 to 200, ResNet prediction models' RMSE and MAPE decrease correspondingly but models' regret increase. This phenomenon means that the hybrid loss improves the models' prediction accuracy at the cost of reducing decision accuracy. {\rv When $\epsilon$ is 1, the regret of the ResNet prediction model is similar to that with $\epsilon$ as 25, but the RMSE and MAPE have doubled than those with $\epsilon$ as 25. In contrast, when $\epsilon$ is above 100, the regret of the ResNet prediction model is almost 3.7 times the value of that with $\epsilon$ as 25 but is still 34 \% of that under MSE-based loss function (only considering $L_{MSE}$). Compared to MSE, the proposed hybrid loss can reduce decision and prediction errors with lower RMSE and lower regret even with the large value of $\epsilon$, which verifies its superiority and robustness.} 

{\rv Compared} with the single loss design, MSE features low RMSE and MAPE but high regret, while the surrogate regret model features high RMSE and MAPE but low regret. Though hybrid loss models endure a little higher regret than the surrogate regret model, they reduce the RMSE and MAPE of single loss models. 

\paragraph{Errors in different time intervals} 
When $\epsilon$ takes the value of 25, the RMSE of MSE and hybrid loss models are similar, which are lower than the surrogate regret model. Then we further investigate hourly prediction errors.

\begin{figure}[ht]
  \centering
  \includegraphics[scale=1]{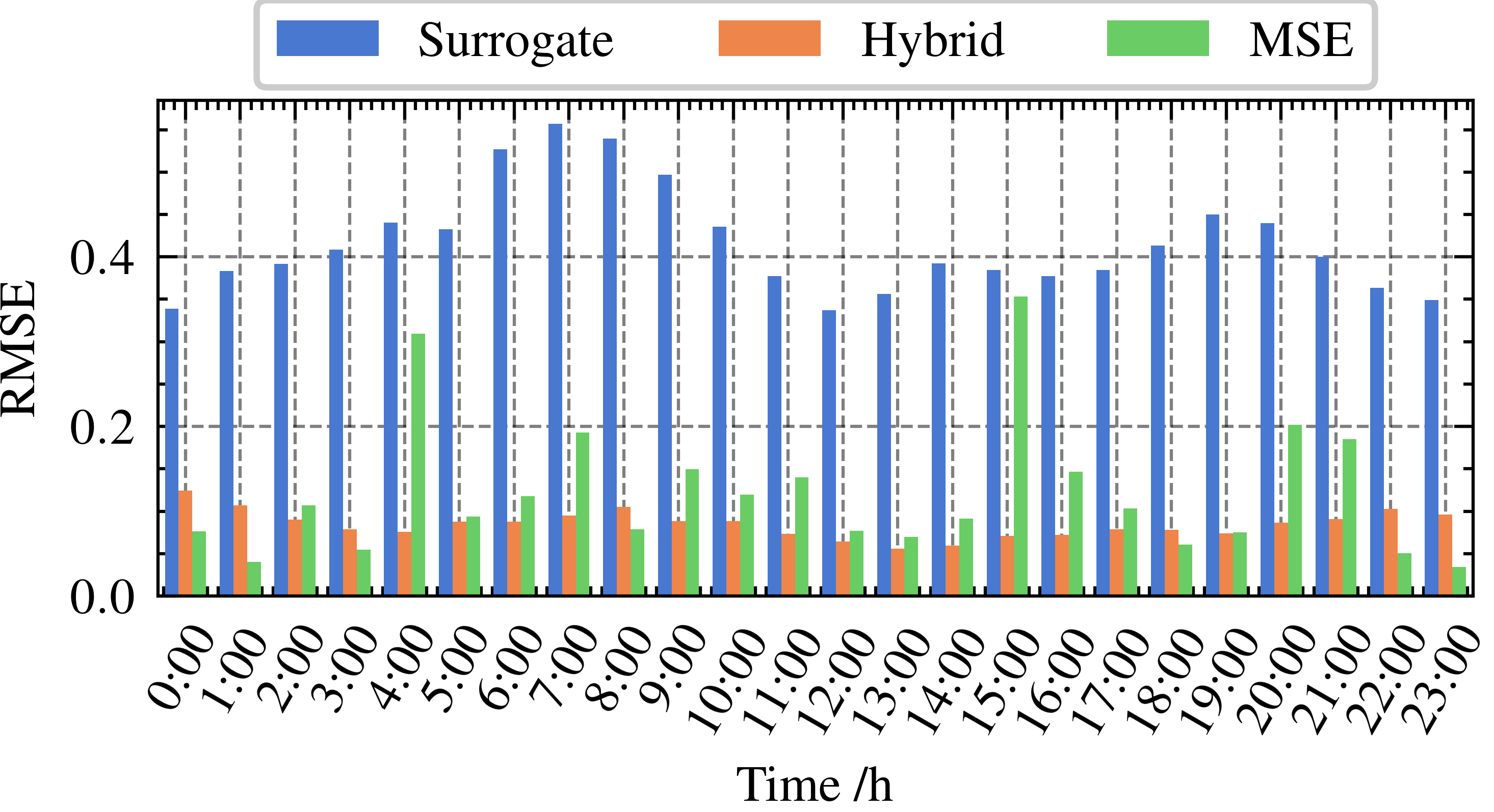}
  \caption{RMSE in different time intervals of different loss function by ResNet model.}
  \label{interval_comparison_Res}
\end{figure}

As shown in Fig. \ref{interval_comparison_Res}, though the hourly prediction errors of surrogate regret are higher than those of MSE and hybrid loss, surrogate regret embodies the lowest decision errors. The daily time distribution of prediction errors under MSE embodies significant variation with high prediction errors in the afternoon (12, 13, and 16), leading to wrong ESS decisions. In contrast, the daily time distribution of prediction errors under hybrid loss embodies low variation and time-stable prediction error.

\paragraph{Results analysis}
Numerical experiments in the ResNet model help further capture the impacts of hybrid loss in large representational capacity prediction models. i) The model parameter updating directions from prediction and decision errors are inconsistent. ii) The gradients from surrogate regret tend to flatten the time distribution of prediction errors, which leads to more accurate decisions. This can be interpreted as the predictable case, where electricity price shows regularity and can be predicted accurately. In this case, the hybrid loss can help reduce prediction errors at the cost of increasing some decision errors {\rv compared} to surrogate regret.

\section{Conclusion}
This paper proposes a decision-focused electricity price prediction approach for ESS arbitrage by considering the reverse impact of downstream optimization models. Based on the surrogate regret and MSE, a hybrid loss is utilized to measure the combined decision and prediction errors. The hybrid SGD learning method is then proposed for training prediction models, including twice back-propagation and once updating. Based on the PJM dataset, numerical experiments indicate that the proposed approach can predict the price changing trend accurately and improve the decision accuracy efficiently by flattening the daily time distribution of prediction errors. Compared to MSE-based models, the decision-focused approach for electricity prediction can bring more economic benefits and reduce prediction errors.

% \section*{Acknowledgments}

\appendices

\ifCLASSOPTIONcaptionsoff
  \newpage
\fi

\begin{IEEEbiography}[{\includegraphics[width=1in,height=1.25in,clip,keepaspectratio]{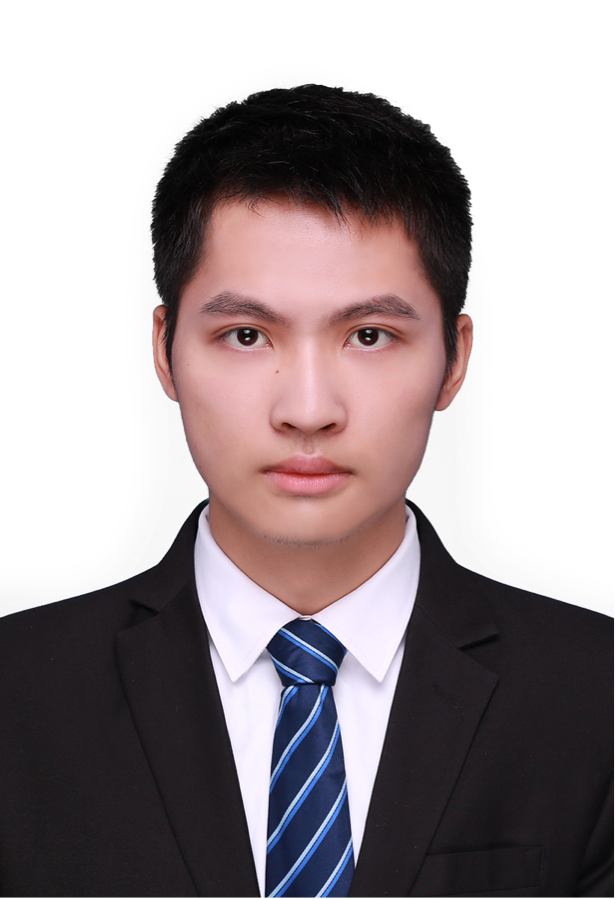}}]{Linwei Sang}(S'20) 
  received the M.S. degree from the School of Electric Engineering, Southeast University, China in 2021. 
  
  He is currently pursuing the Ph.D. degree in the Tsinghua-Berkeley Shenzhen Institute, Tsinghua University, Shenzhen, China. His research includes machine learning application in smart grid, the control of the distributed energy, and demand side resource management. 
\end{IEEEbiography}

\begin{IEEEbiography}[{\includegraphics[width=1in,height=1.25in,clip,keepaspectratio]{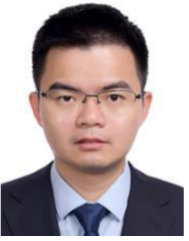}}]{Yinliang Xu}(SM'19)
  received the B.S. and M.S. degrees in control science and engineering from the Harbin Institute of Technology, Harbin, China, in 2007 and 2009, respectively, and the Ph.D. degree in electrical and computer engineering from New Mexico State University, Las Cruces, NM, USA, in 2013.
  
  He is currently an Associate Professor with Tsinghua-Berkeley Shenzhen Institute, Tsinghua Shenzhen International Graduate School, Tsinghua University, Beijing, China. His research interests include distributed control and optimization of power systems, renewable energy integration, and microgrid modeling and control. 
\end{IEEEbiography}

\begin{IEEEbiographynophoto}{Huan Long}(M'15)
  received the B.Eng. degree from Huazhong University of Science and Technology, Wuhan, China, in 2013, and the Ph.D. degree from the City University of Hong Kong, Hong Kong, in 2017. 
  
  She is currently an Associate Professor with the School of Electrical Engineering, Southeast University, Nanjing, China. Her research fields include artificial intelligence applied in modeling, optimizing, monitoring the renewable energy system and power system. 
\end{IEEEbiographynophoto}

\begin{IEEEbiographynophoto}{Qinran Hu}(SM'21) received the B.S.
  degree from Chien-Shiung Wu College, Southeast University, Nanjing, China, in 2010, and
  the M.S. and Ph.D. degrees from the University
  of Tennessee, Knoxville, TN, USA, in 2013 and
  2015, respectively, all in electrical engineering.
  
  He was a Postdoctoral Fellow with Harvard
  University, Cambridge, MA, USA, from 2015 to
  2018. He joined the School of Electrical Engineering, Southeast University, in October 2018.
  His research interests include power system optimization, demand aggregation, and virtual power plant.
\end{IEEEbiographynophoto}

\begin{IEEEbiographynophoto}{Hongbin Sun}(Fellow, IEEE) received his double B.S. degrees from Tsinghua University in 1992, the Ph.D from Dept. of E.E., Tsinghua University in 1996. 
  
  He is now ChangJiang Scholar Chair professor and the director of energy management and control research center, Tsinghua University. He also serves as the editor of the IEEE TSG, associate editor of IET RPG, and member of the Editorial Board of four international journals and several Chinese journals. His technical areas include electric power system operation and control with specific interests on the Energy Management System, Automatic Voltage Control, and Energy System Integration.
\end{IEEEbiographynophoto}

\clearpage

\end{document}